\documentclass[conference]{IEEEtran}
\usepackage{fancyhdr}
\IEEEoverridecommandlockouts

\usepackage{cite}
\usepackage{amsmath,amssymb,amsfonts}
\usepackage{algorithmic}
\usepackage{algorithm}
\usepackage{graphicx}
\usepackage{textcomp}
\usepackage{xcolor}
\usepackage{booktabs}
\usepackage{subcaption}
\usepackage{url}
\usepackage{xurl}
\usepackage[capitalize]{cleveref}
\usepackage{multirow}
\usepackage{makecell}

\def\BibTeX{{\rm B\kern-.05em{\sc i\kern-.025em b}\kern-.08em
    T\kern-.1667em\lower.7ex\hbox{E}\kern-.125emX}}
\begin{document}

\title{TIDE: Temporal Incremental Draft Engine for Self-Improving LLM Inference}

\author{
\IEEEauthorblockN{Jiyoung Park, Hankyu Jang, Changseok Song, and Wookeun Jung\textsuperscript{*}\thanks{\textsuperscript{*}Corresponding author.}}
\IEEEauthorblockA{\textit{Moreh, Inc.}\\
Seoul, South Korea \\
\{jiyoung.park, hankyu.jang, changseok.song, wookeun.jung\}@moreh.io}
}

\maketitle

\thispagestyle{fancy} 
\lhead{} 
\rhead{} 
\chead{} 
\lfoot{\footnotesize{ 
SC26, November 15-20, 2026, Chicago, Illinois, USA 
\newline 979-8-3195-4789-7/26/\$31.00 \copyright 2026 IEEE}}
\rfoot{} 
\cfoot{} 
\renewcommand{\headrulewidth}{0pt} 
\renewcommand{\footrulewidth}{0pt}

\begin{abstract}
Speculative decoding can substantially accelerate LLM inference, but realizing its benefits in practice is challenging due to evolving workloads. We present TIDE (Temporal Incremental Draft Engine), a serving-engine-native framework that integrates online draft adaptation directly into high-performance LLM inference systems. TIDE reuses target model's intermediate hidden states generated during inference as training signals for draft adaptation, thereby avoiding additional target model computation and serving-time overhead. It employs adaptive runtime control to activate speculation and draft model training only when beneficial. TIDE exploits heterogeneous clusters by mapping inference and training to appropriate GPU classes. Across diverse real-world workloads, TIDE achieves up to 1.66× throughput over no-speculation baselines while recovering performance on misaligned workloads where static draft models degrade throughput. TIDE also reduces training time by up to 3.02× and storage requirements by 24× compared to existing draft training approaches, and improves system throughput by up to 1.22× on heterogeneous GPU clusters.
\end{abstract}


\section{Introduction}

Large language models (LLMs) increasingly achieve state-of-the-art performance on reasoning-intensive tasks, such as mathematics, code generation, and scientific problem solving, by scaling test-time computation---for example, by generating longer reasoning traces or sampling multiple candidate solutions \cite{snell2024scalingllmtesttimecompute, muennighoff2025s1simpletesttimescaling}. As these approaches increase the amount of autoregressive generation per request, decoding efficiency has become a central bottleneck for deploying reasoning-oriented LLMs in production systems \cite{davies2025liminalexploringfrontiersllm}.

\begin{figure}[t]
  \centering
  \includegraphics[width=\columnwidth]{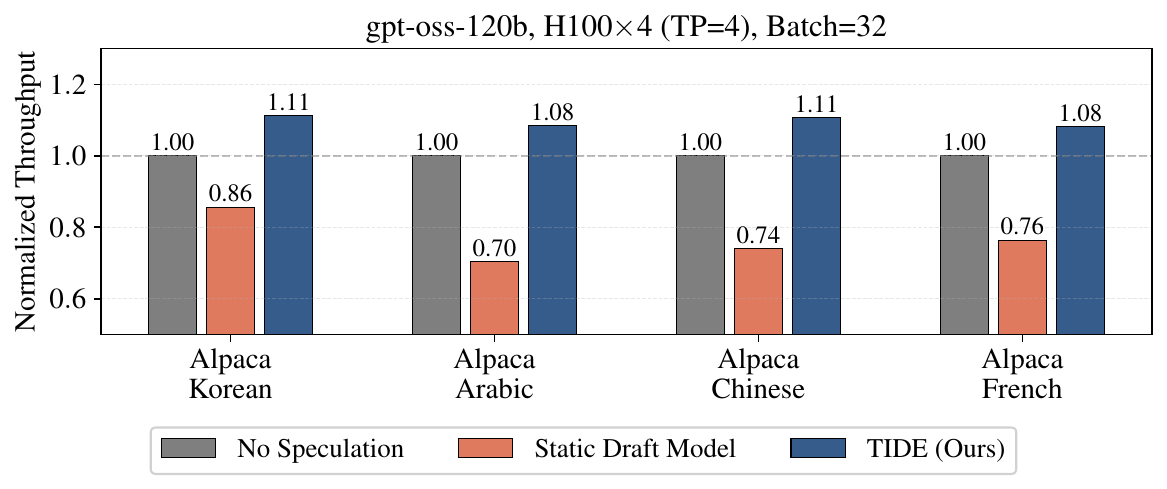}
  \caption{Normalized throughput of speculative decoding with a static draft model versus TIDE across four non-English Alpaca datasets. The static draft model, trained on English data, degrades throughput below the no-speculation baseline on all workloads. TIDE recovers and exceeds baseline throughput by continuously adapting the draft model to the current workload.}
  \label{fig:spec_decode_comparison}
\end{figure}

Speculative decoding is one of the most effective techniques for accelerating LLM inference. By allowing a lightweight draft model to propose multiple draft tokens that are then verified in parallel by a target model, speculative decoding can significantly improve throughput and latency \cite{leviathan2023fastinferencetransformersspeculative, chen2023acceleratinglargelanguagemodel}. However, these gains depend on \emph{draft--target alignment}---the degree to which the draft model's next-token predictions approximate those of the target model \cite{zhou2024distillspecimprovingspeculativedecoding,hong2025trainingdomaindraftmodels}. When alignment is poor, the target model rejects most drafted tokens, and speculative decoding can even degrade overall serving performance (see \cref{fig:spec_decode_comparison}). Indeed, while major inference service providers support speculative decoding, the feature is typically disabled by default, with providers cautioning that it does not guarantee speedups and may even increase latency or reduce throughput depending on the workload and model configuration \cite{fireworks2024specdec, togetherai2025specdec, baseten2024specdec}.

A fundamental challenge is that draft--target alignment is inherently workload-dependent. In production LLM services, inference workloads evolve continuously as user behavior changes, application logic is updated, and prompt templates are modified. While workload distributions are globally non-stationary, prior studies show that they exhibit strong short-term temporal locality, with recent inference history remaining predictive of near-future requests \cite{wang2024burstgpt, gim2024promptcachemodularattention, zheng2024lmsyschat1mlargescalerealworldllm, xiang2025servegenworkloadcharacterizationgeneration}. This suggests that alignment can, in principle, be preserved by continuously and rapidly adapting to recent inference behavior.

Recent work has explored this opportunity by adapting draft models online using inference-time signals, for example via online distillation from target model corrections or logits \cite{liu2024onlinespeculativedecoding}. While these approaches demonstrate that alignment can be recovered under distribution shift, they primarily focus on the learning algorithm itself. It remains an open question whether online draft training can be integrated into high-performance inference engines while delivering sustained end-to-end throughput improvements \cite{zhou2024distillspecimprovingspeculativedecoding, yan2025decodingspeculativedecoding}.

In practice, addressing this question requires careful coordination between online draft model training and inference serving. Online draft-model training must minimize interference with latency-critical inference, operate under realistic resource constraints, and be triggered only when beneficial. Because the performance impact of speculative decoding varies across workload phases,
indiscriminate use of speculative decoding or draft model training is often unnecessary and can even be counterproductive \cite{Huang_2025,huang2025specdecboostingspeculativedecoding,hou2025banditspecadaptivespeculativedecoding}. Effective deployment therefore requires dynamic runtime control over \emph{when} to speculate and \emph{when} to train, based solely on signals observable during inference serving.

To address these challenges, we introduce \textbf{Temporal Incremental Draft Engine (TIDE)}, a serving-engine-native framework for adaptive speculative decoding under evolving workloads. Rather than treating draft adaptation as an isolated learning problem, TIDE jointly manages training signal collection, draft model updates, and speculative decoding decisions entirely within the inference serving engine.

TIDE exploits short-term temporal locality in workload distributions by incrementally training the draft model on recent inference data, while dynamically controlling when speculative decoding and draft model training are beneficial. Crucially, since draft model training requires the target model's intermediate hidden states for the corresponding input tokens as supervision signals, a na\"ive approach would need to reload the large target model and recompute forward passes solely for training---an operation that would dominate training cost given the target model's scale. TIDE avoids this recomputation by reusing intermediate hidden states already produced during target model inference. To deliver these training signals to the separate draft-model training engine, TIDE copies them from GPU memory to a host buffer and writes the buffered data to shared storage. It performs both operations asynchronously and overlaps them with subsequent target-model computation, keeping training-signal extraction off the serving critical path and allowing the draft model to adapt fast enough to keep pace with shifting workload distributions.

This separation also enables efficient deployment on heterogeneous hardware. Heterogeneous accelerators are receiving increasing attention in LLM serving \cite{mei2025helix, griggs2024melange, zhang2025cauchy}. As an additional industry example, NVIDIA's Vera Rubin platform pairs GPUs with specialized low-latency processors (Groq 3 LPX) to serve different phases of inference on purpose-built hardware \cite{nvidia2026groq3lpx}. TIDE embraces this direction by enabling inference and draft model training to be assigned to different device types according to their computational characteristics. In our evaluation, we assign inference to NVIDIA H100 GPUs, which are optimized for low-latency serving, and dedicate AMD Instinct MI250 GPUs to draft model adaptation. Compared to a baseline that allocates both GPU types to inference with a static draft model, TIDE achieves higher overall system throughput through continuous draft--target alignment improvement. As heterogeneous clusters become the industry standard rather than the exception, frameworks like TIDE that effectively leverage diverse hardware for both inference serving and draft model training are essential for maximizing system-wide efficiency.

In summary, our main contributions are:
\begin{itemize}
  \item We propose TIDE, a serving-engine-native framework for adaptive speculative decoding that incrementally maintains draft--target alignment under inference workloads that are globally non-stationary yet exhibit strong short-term temporal locality.
  \item We generate training signals by reusing intermediate hidden states computed during inference and asynchronously delivering them to the draft-model training engine through shared storage, overlapping the required data movement with subsequent target-model computation to keep it off the serving critical path.
  \item We introduce adaptive runtime control mechanisms that determine when to speculate and when to train, enabling speculative decoding only when it yields throughput gains and triggering draft model training only when alignment degradation is detected.
  \item We demonstrate effective heterogeneous GPU utilization by decoupling inference and training, assigning each workload to the hardware best suited for it: latency-sensitive inference on NVIDIA H100 GPUs and throughput-oriented training on AMD MI250 GPUs.
  \item We implement a complete TIDE prototype and show consistent system-level throughput improvements across diverse real-world workload patterns.
\end{itemize}

\section{Background}
\begin{figure}[!t]
  \vskip 0.2in
  \begin{center}
    \centerline{\includegraphics[width=\columnwidth]{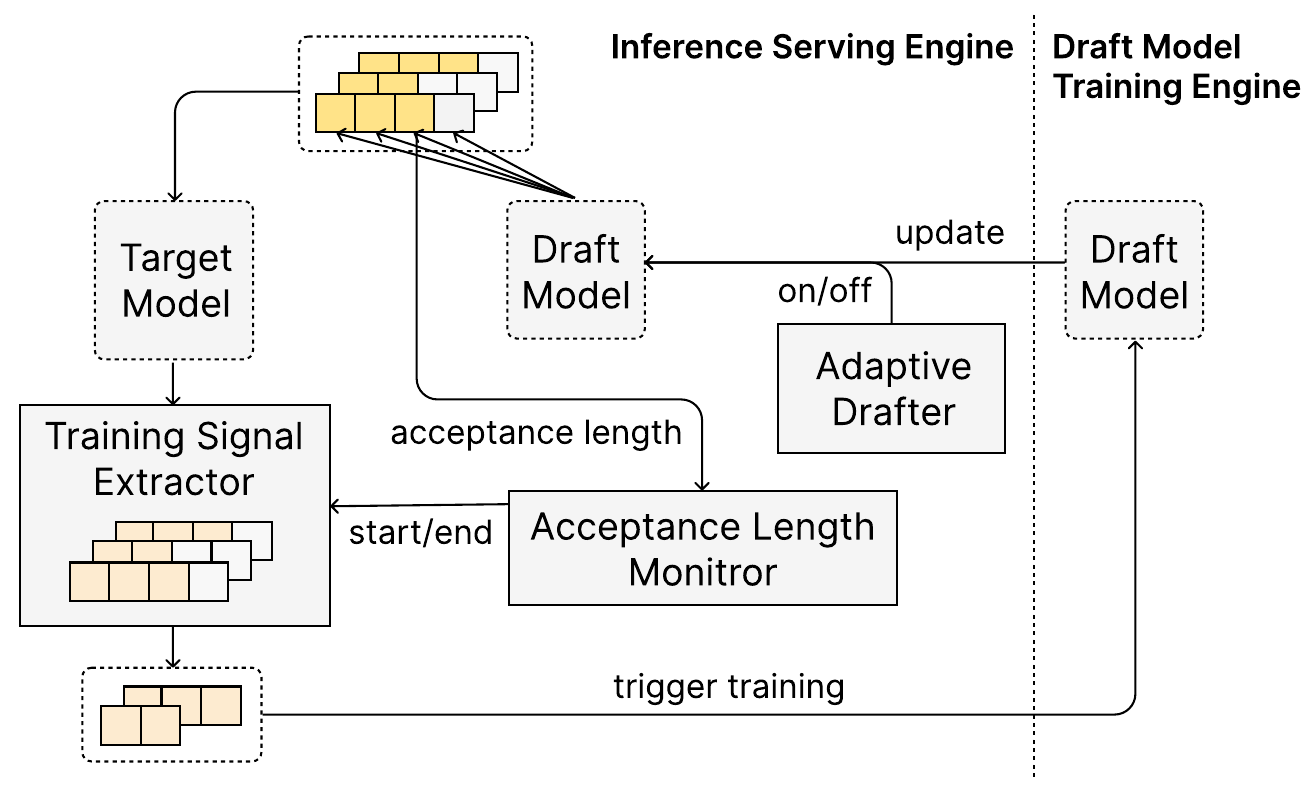}}
    \caption{Overview of TIDE architecture and workflow.}
    \label{fig:overview}
  \end{center}
\end{figure}
\subsection{Speculative Decoding in Dynamic Workloads}
Large language models (LLMs) generate text autoregressively: each token is produced sequentially, with every generation step depending on all previously generated tokens. Because a single decoding step must read the entire model's parameters from GPU memory yet performs relatively little arithmetic, LLM decoding is typically memory-bandwidth bound \cite{kwon2023efficientmemorymanagementlarge,leviathan2023fastinferencetransformersspeculative}.

\begin{figure}[!t]
  \vskip 0.2in
  \begin{center}
    \centerline{\includegraphics[width=\columnwidth]{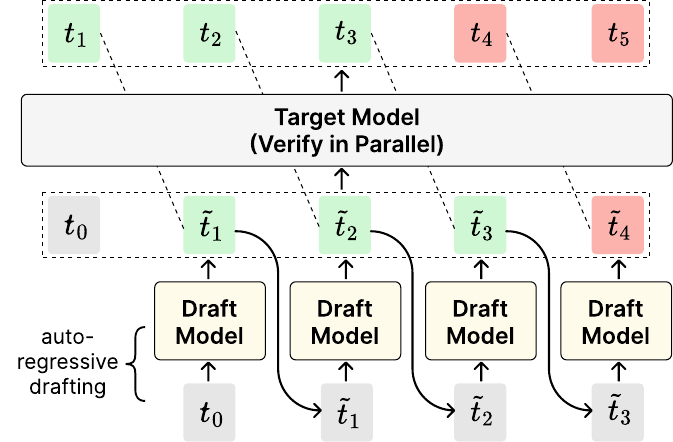}}
  \caption{Overview of speculative decoding with $\gamma = 4$. The draft model autoregressively generates four candidate tokens $\tilde{t}_{1}, \ldots, \tilde{t}_{4}$ from a prefix token $t_0$. The target model verifies all candidates in a single parallel forward pass, producing $t_1, \ldots, t_5$. In this example, the first three draft tokens match the target model's outputs ($\tilde{t}_i = t_i$ for $i = 1, 2, 3$; \textcolor{green!60!black}{green}), while $\tilde{t}_4 \neq t_4$ causes rejection from the fourth token onward (\textcolor{red!70!black}{red}). Had $\tilde{t}_4$ also been accepted, $t_5$ would be obtained as a bonus token at no additional cost.}
    \label{fig:speculative_decoding}
  \end{center}
\end{figure}

Speculative decoding exploits this underutilization. As illustrated in \cref{fig:speculative_decoding}, a smaller draft model $q$ autoregressively generates $\gamma$ candidate tokens, which the target model $p$ then verifies in parallel in a single forward pass. Because the target model's decoding step is memory-bound, verifying a small batch of candidates costs roughly the same wall-clock time as generating a single token, yielding a speedup whenever the draft model's predictions are sufficiently accurate. Leviathan et al.~\cite{leviathan2023fastinferencetransformersspeculative} derived the theoretical speedup of speculative decoding relative to vanilla autoregressive decoding:
\begin{equation}
\label{equation:theoretical_speedup}
Theoretical\ Speedup = \frac{1 - \alpha^{\gamma+1}}{(1 - \alpha)(c\gamma + 1)}
\end{equation}
where $\alpha$ is the token acceptance rate---the probability that a candidate token drawn from the draft model is also consistent with the target model's output distribution, $\gamma$ is the number of candidate tokens, and $c$ is the ratio between draft and target model latency.

This formula reveals three critical conditions for performance improvements: (1) \textbf{high acceptance rate} requiring strong draft--target alignment, (2) \textbf{memory-bound regime} where verification latency equals single-token decoding latency, and (3) \textbf{minimal draft overhead} achievable through lightweight draft model architectures. Note that condition~(2) weakens as the serving batch size grows: with more concurrent requests, verification becomes compute-bound and processing extra candidate tokens adds measurable latency, reducing the benefit of speculative decoding.

Prior work addresses these conditions individually. Online training approaches~\cite{liu2024onlinespeculativedecoding,ramakrishnan2025omnidraft,qian2026whendraftsevolve,kumar2026testtimespeculation} improve draft--target alignment through continuous fine-tuning using inference-time signals. This is motivated by the observation that the acceptance rate is inherently workload-dependent: as the domain or pattern of user requests shifts over time, a previously well-aligned draft model may become inaccurate, degrading speculative decoding performance. Adaptive systems~\cite{Huang_2025,huang2025specdecboostingspeculativedecoding,hou2025banditspecadaptivespeculativedecoding,liu2024turbospec,zhang2025draftmodelknows,agrawal2024adaedl} focus on runtime control by dynamically adjusting the number of speculative tokens or disabling speculative decoding based on system load and acceptance rates. Lightweight draft architectures~\cite{cai2024medusasimplellminference,li2025eagle3scalinginferenceacceleration,ankner2024hydra,zhang2024recurrentdrafter} reduce draft model overhead by simplifying model structure or reusing intermediate representations.

However, these techniques have only been evaluated individually under controlled experimental settings, without demonstrating their benefits when integrated into a complete serving system. In particular, there is limited investigation into how these techniques should be applied in practice to improve end-to-end throughput in real LLM serving environments.

\subsection{EAGLE: Hidden-State-Based Draft Models}

Recent draft model architectures such as EAGLE~\cite{li2025eagle3scalinginferenceacceleration} achieve minimal overhead by reusing the target model's own internal representations. In a Transformer-based LLM, each input token is transformed through a stack of decoder layers, producing hidden states---high-dimensional vectors that encode increasingly abstract semantic information at each layer. The EAGLE-family draft model consists of only a single decoder layer and a lightweight output projection (LM head), and takes concatenated hidden states from multiple decoder layers (e.g., early, middle, and late layers) of the target model as input. Because these hidden states are already computed during normal inference, they can be directly reused as both the draft model's input and its training signal, requiring no additional target model computation.

\subsection{Heterogeneous Systems for LLM Inference}

Modern datacenters inevitably consist of heterogeneous GPU clusters, as new accelerator generations are adopted incrementally over time. Prior work has primarily exploited such heterogeneity for inference-only execution, either by assigning different models to different GPUs based on coarse-grained criteria (e.g., model size or latency sensitivity) \cite{duan2024muxserveflexiblespatialtemporalmultiplexing, xiang2025aegaeon}, or by distributing different inference phases of a single model across heterogeneous devices, such as disaggregated prefill and decoding \cite{zhong2024distservedisaggregatingprefilldecoding, jiang2025hexgen2, chen2025multivendorpd}. In these systems, heterogeneity is handled via static or periodically updated placement policies derived from offline profiling.

In contrast, TIDE considers heterogeneity in a setting where inference is interleaved with online draft model adaptation. Rather than optimizing inference alone, TIDE decouples inference serving and draft training and assigns them to different GPU classes based on their respective performance and cost characteristics. This enables heterogeneous clusters to efficiently support adaptive speculative decoding, a dimension not addressed by prior heterogeneous LLM serving systems.

\section{TIDE Architecture}
\subsection{System Overview}
\begin{figure}[!t]
  \vskip 0.2in
  \begin{center}
    \centerline{\includegraphics[width=\columnwidth]{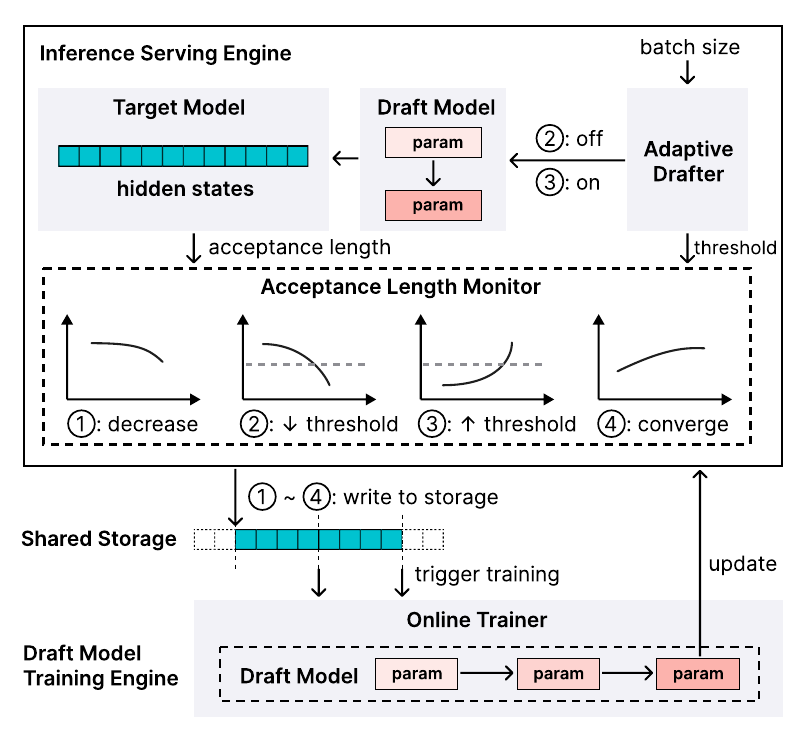}}
    \caption{Detailed TIDE architecture and workflow. The system monitors acceptance length to adaptively enable/disable speculative decoding and selectively trigger training signal collection based on workload changes.}
    \label{fig:architecture}
  \end{center}
\end{figure}
As illustrated in \cref{fig:overview}, TIDE consists of the Inference Serving Engine with its Adaptive Drafter, Acceptance Length Monitor, and Training Signal Extractor, and the Draft Model Training Engine, working collaboratively to maintain high inference throughput while continuously adapting the draft model to evolving workloads.

\textbf{Inference Serving Engine.} The Inference Serving Engine handles incoming requests using speculative decoding with adaptive control. During each inference iteration, the draft model generates candidate tokens, which are then verified by the target model in parallel. 
This verification process produces two critical outputs: per-request acceptance length, the number of draft tokens accepted by the target model for each request, and draft model training signals, consisting of target model's intermediate hidden states generated during prefill and verification phases for input and output tokens.

Based on these outputs, two decision components operate as illustrated in \cref{fig:architecture}: The Adaptive Drafter monitors batch size and acceptance length to determine whether to enable or disable speculative decoding, ensuring it is only applied when beneficial for performance (Section~\ref{subsection:adaptive_speculative_decoding_control}). The Training Signal Extractor decides whether to store training signals in shared storage, minimizing storage and training overhead. It halts extraction when the acceptance length improvement plateaus, indicating the draft model has sufficiently adapted to the current workload distribution, and resumes extraction upon detecting distribution shift (Section~\ref{subsection:selective_draft_model_training}).

\textbf{Draft Model Training Engine.} The Draft Model Training Engine operates asynchronously on separate GPU resources independent from the Inference Serving Engine. It continuously monitors shared storage for accumulated training signals. When the signals exceed a predefined threshold, it triggers a training cycle, loading the current draft model checkpoint and fine-tuning it using the accumulated training signals. After training converges, the updated draft model is evaluated and deployed to the Inference Serving Engine only if it demonstrates improved acceptance length.

\subsection{Serving-Time Training Signal Extraction}
\begin{figure}[t]
  \centering
  \includegraphics[width=\columnwidth]{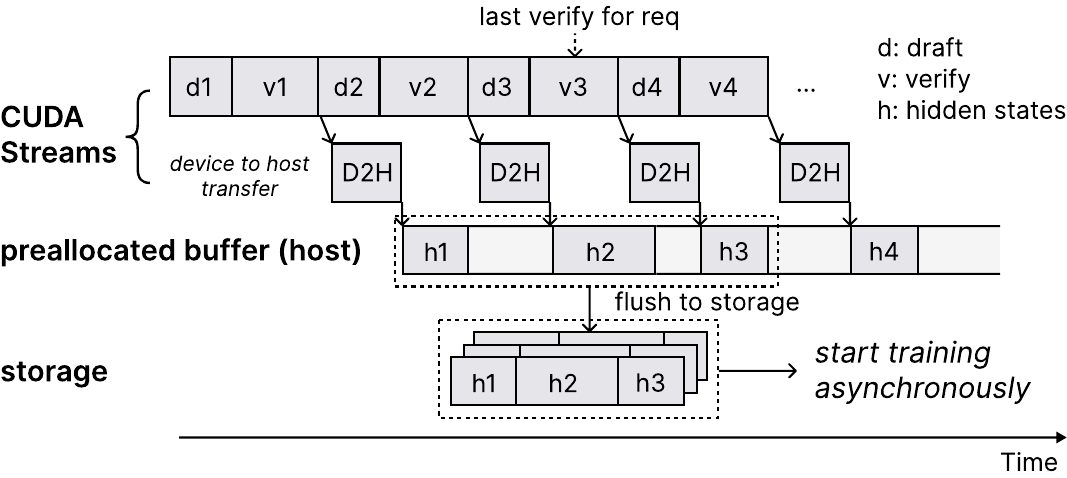}
  \caption{TIDE's asynchronous adaptation pipeline. Device-to-host transfer and buffered storage writes overlap with subsequent GPU verification (top), keeping training-signal extraction off the serving critical path, while draft model training proceeds concurrently on separate resources (bottom).}
  \label{fig:async_training}
\end{figure}
TIDE employs EAGLE-3~\cite{li2025eagle3scalinginferenceacceleration}, a state-of-the-art speculative decoding technique, to build a draft model consisting of a single decoder layer and an LM head. The EAGLE-3 draft model predicts the next token based on the target model's intermediate hidden states, rather than directly from the input prompt as in conventional LLMs. Specifically, EAGLE-3 uses concatenated hidden states from decoder layers at low, middle, and high positions of the target model, leading to richer semantic information for next token predictions. These hidden states are computed during the target model's prefill for input prompts and decoding (or verification if speculative decoding is enabled) phases, making them natural byproducts of normal inference operations that can be directly used as training signals for the EAGLE-3 draft model.

To leverage this opportunity for efficient training signal collection, we implement intermediate hidden state extraction on top of SGLang~\cite{zheng2024sglangefficientexecutionstructured} and vLLM~\cite{kwon2023efficientmemorymanagementlarge}, making our approach applicable to any transformer-based architecture. Extraction adds two serving-side operations: (1) the hidden states for accepted tokens are concatenated and copied from device memory to a pre-allocated host buffer, and (2) the buffer is flushed to shared storage when full. As illustrated in \cref{fig:async_training}, both operations execute asynchronously and overlap with the next verification step's compute kernels, keeping training-signal extraction off the serving critical path despite consuming memory, interconnect, and storage bandwidth. Draft model training likewise consumes separate compute resources but proceeds asynchronously from inference serving.

\subsection{Online Draft Model Adaptation}
Traditional draft model training approaches \cite{specforge2025, hong2025trainingdomaindraftmodels} require running target model inference again during the training process to generate training data. This leads to two significant limitations: (1) substantially increased training time as analyzed in \cref{subsection:training_efficiency_comparison}, and (2) excessive GPU memory consumption from loading both the target and draft models simultaneously, constraining the resources available for training. 

TIDE addresses these limitations by directly utilizing training signals extracted from the inference serving engine, eliminating redundant computation. The target model's intermediate hidden states—used as training data for the draft model—are already generated during inference serving, meaning only the compact draft model (a single decoder layer with an LM head) needs to be loaded onto the device for training, regardless of the target model's size. This enables TIDE to scale to arbitrarily large target models without increasing training resource requirements.

As illustrated in \cref{fig:async_training}, draft model training proceeds asynchronously from inference serving. The draft model's compact architecture, combined with multi-device parallelization using PyTorch FSDP~\cite{zhao2023pytorchfsdpexperiencesscaling}, enables real-time adaptation even with relatively modest resource allocation dedicated to training.

\section{Adaptive Control and Runtime Optimization}
\subsection{Adaptive Speculative Decoding Control}
\label{subsection:adaptive_speculative_decoding_control}
\begin{figure*}[t]
  \centering
  \includegraphics[width=\textwidth]{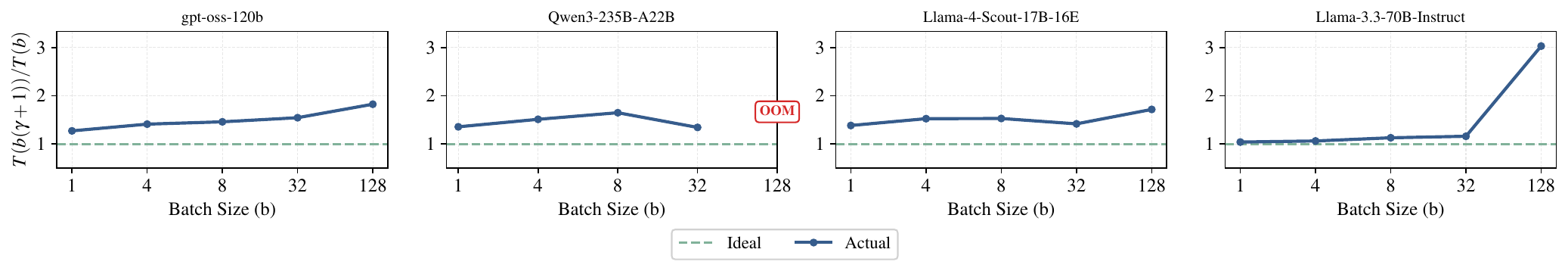}
\caption{Ratio of verification latency $T(b(\gamma+1))$ for $\gamma=3$ candidate tokens to single-token decoding latency $T(b)$ across different batch sizes. If decoding is completely memory-bound, this ratio would be 1.0 (ideal case shown by dashed line). The ratio exhibits substantial variance depending on both batch size and model architecture. Since the ratio exceeds 1.0 in most configurations, the effective speedup from speculative decoding is reduced accordingly, underscoring the need for adaptive control that selectively enables speculation only when conditions are favorable.}
  \label{fig:verification_latency_ratio}
\end{figure*}
Speculative decoding fundamentally relies on the memory-bound nature of standard autoregressive decoding to achieve speedup. Even when draft--target alignment is strong and draft model overhead is minimal, performance gains can be limited by batch size. As concurrent requests increase, target model decoding shifts from memory-bound to compute-bound, increasing verification overhead and diminishing the benefits of speculative decoding.

This section presents a performance model that estimates the speedup of speculative decoding as a function of batch size, enabling TIDE to adaptively apply speculative decoding only when it provides measurable benefits. We derive a practical speedup formula that accounts for how batch size affects speculative decoding performance.

In each speculation step, the draft model generates $\gamma$ candidate tokens, of which an average fraction $\alpha$ is accepted during target model verification. The expected acceptance length is~\cite{leviathan2023fastinferencetransformersspeculative,chen2023acceleratinglargelanguagemodel}:
\vspace{2pt}
\begin{equation}
E[\ell] = \frac{1 - \alpha^{\gamma+1}}{1 - \alpha}
\end{equation}
\vspace{2pt}
Let $D(n)$ and $T(n)$ denote the latency for the draft model and the target model to decode $n$ tokens in parallel, respectively. When processing $b$ concurrent requests in a batch, the per-token latency with speculative decoding becomes:
\vspace{2pt}
\begin{equation}
SD(b) = \frac{\gamma D(b) + T(b(\gamma+1))}{E[\ell]}
\end{equation}
\vspace{2pt}

where $\gamma D(b)$ represents the latency of generating $\gamma$ candidate tokens by the draft model, and $T(b(\gamma+1))$ represents the latency of verifying these $\gamma$ tokens plus one bonus token from the previous step by the target model.

The speedup relative to standard autoregressive decoding is:
\vspace{2pt}
\begin{align}
\label{eq:speedup}
Speedup &= \frac{T(b)}{SD(b)} \nonumber \\
&= \frac{1 - \alpha^{\gamma+1}}{(1 - \alpha)\left( \frac{D(b)}{T(b)}\gamma+\frac{T(b(\gamma+1))}{T(b)}\right)}
\end{align}
\vspace{2pt}

The theoretical speedup formula in Equation~\ref{equation:theoretical_speedup} is derived by substituting $c = D(b) / T(b)$ and approximating $T(b(\gamma+1)) / T(b) \approx 1$, which holds only in the memory-bound regime. However, this assumption often fails to predict actual performance. As batch size increases and the standard decoding becomes compute-bound, the ratio $\beta(b) = T(b(\gamma+1)) / T(b)$ grows significantly (see \cref{fig:verification_latency_ratio}).

Moreover, the coefficient $c = D(b) / T(b)$ is not static in practice. Modern speculative decoding architectures such as EAGLE-3 \cite{li2025eagle3scalinginferenceacceleration} consist of a single decoder layer and an LM head, making the draft model's computational cost negligible compared to the target model. However, in implementations such as vLLM \cite{kwon2023efficientmemorymanagementlarge} and SGLang \cite{zheng2024sglangefficientexecutionstructured}, the draft model latency is dominated by CPU overhead including kernel launch overhead, which is independent of batch size.
Therefore, we can approximate $D(b) \approx D_0$ as a static value, while $T(b)$ varies with batch size, causing $c(b) = D_0 / T(b)$ to decrease as batch size increases. 

Incorporating these observations, we obtain a simplified practical speedup formula:
\vspace{2pt}
\begin{equation}
\label{eq:practical_speedup}
Practical~Speedup = \frac{1 - \alpha^{\gamma+1}}{(1 - \alpha)\left(c(b)\gamma +\beta(b)\right)}
\end{equation}
\vspace{2pt}
Given $T(n)$ and $D_0$, we can determine the minimum acceptance rate required for speculative decoding to provide performance gains at a given batch size. Since $T(n)$ and $D_0$ vary depending on the model architecture, parallelization strategy, and system environment, they must be empirically profiled. TIDE's Adaptive Drafter profiles $T(n)$ and $D_0$ during system initialization by measuring latencies across different batch sizes. These profiled values are used to estimate speedup in real-time according to Equation~\ref{eq:practical_speedup}, enabling TIDE to adaptively enable or disable speculative decoding based on current batch size and acceptance rates, thereby maintaining peak inference throughput.

\begin{algorithm}[!ht]
  \caption{Selective Training Control}
  \label{alg:adaptive_threshold}
  \small
  \begin{algorithmic}
    \STATE {\bfseries Input:} $\lambda_{\text{short}}, \lambda_{\text{long}}, \epsilon, N_{\text{init}}, N_{\text{threshold}}$
    \STATE $\text{collection\_enabled} \gets \text{True}$, $M_{\text{draft}} \gets $ initial draft model
    \STATE
    \STATE Measure acceptance rates $\{\alpha_1, \ldots, \alpha_{N_{\text{init}}}\}$ from first $N_{\text{init}}$ requests
    \STATE $\bar{\alpha}_{\text{short}}, \bar{\alpha}_{\text{long}} \gets \frac{1}{N_{\text{init}}} \sum_{i=1}^{N_{\text{init}}} \alpha_i$
    \STATE
    \WHILE{serving requests}
    \STATE Extract target-model intermediate hidden states $h$ and measure acceptance rate $\alpha$
    \STATE $\bar{\alpha}_{\text{short}} \gets \lambda_{\text{short}} \bar{\alpha}_{\text{short}} + (1 - \lambda_{\text{short}}) \alpha$
    \STATE $\bar{\alpha}_{\text{long}} \gets \lambda_{\text{long}} \bar{\alpha}_{\text{long}} + (1 - \lambda_{\text{long}}) \alpha$
    \STATE
    \IF{$\bar{\alpha}_{\text{short}} < \bar{\alpha}_{\text{long}} - \epsilon$}
    \STATE $\text{collection\_enabled} \gets \text{True}$
    \ENDIF
    \STATE
    \IF{$\text{collection\_enabled}$}
    \STATE Store $(h, \alpha)$
    \ENDIF
    \STATE
    \IF{$|\text{stored samples}| \geq N_{\text{threshold}}$}
    \STATE Split into $D_{\text{train}}, D_{\text{eval}}$
    \STATE $M_{\text{new}} \gets \text{train}(M_{\text{draft}}, D_{\text{train}})$
    \STATE $\bar{\alpha}_{\text{train}} \gets $ average $\alpha$ in $D_{\text{train}}$
    \STATE $\bar{\alpha}_{\text{eval}} \gets \text{eval}(M_{\text{new}}, D_{\text{eval}})$
    \IF{$\bar{\alpha}_{\text{eval}} > \bar{\alpha}_{\text{train}}$}
    \STATE $M_{\text{draft}} \gets M_{\text{new}}$
    \ELSIF{$\bar{\alpha}_{\text{eval}} < \bar{\alpha}_{\text{train}}$}
    \STATE $\text{collection\_enabled} \gets \text{False}$
    \ENDIF
    \ENDIF
    \ENDWHILE
  \end{algorithmic}
\end{algorithm}

\subsection{Selective Draft Model Training}
\label{subsection:selective_draft_model_training}
\begin{figure}[t]
  \centering
  \includegraphics[width=1.03\columnwidth]{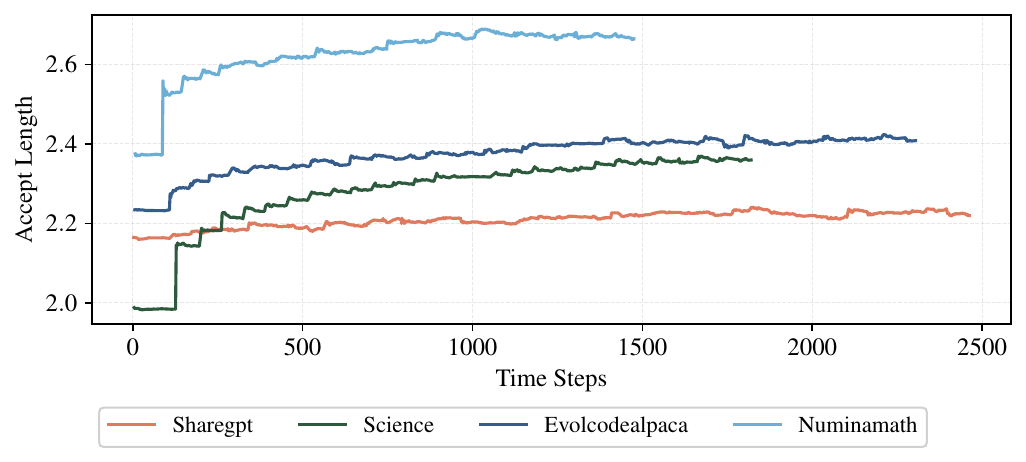}
  \caption{Accept length evolution during draft model training across four datasets using gpt-oss-120b as the target model. Accept length measures the average number of tokens accepted per speculative decoding step. Each time step corresponds to 30 seconds of training time.}
  \label{fig:acc_length_converge}
\end{figure}

Training on every incoming request would waste computational resources and could lead to overfitting on the observed distribution. TIDE employs a selective training strategy that dynamically enables or disables training signal collection based on whether the draft model training is effective, ensuring efficient adaptation while avoiding unnecessary computation once training yields diminishing returns.

As shown in \cref{fig:acc_length_converge}, accept length exhibits a characteristic saturation pattern during draft model training. This saturation behavior motivates TIDE's selective training strategy, which adaptively pauses training when further updates yield minimal acceptance improvements, reducing computational overhead while maintaining draft model quality.

The key challenge is detecting when draft model training is no longer effective—when training on the collected data does not improve acceptance rates. TIDE addresses this by monitoring distribution shifts through acceptance rate patterns and evaluating training effectiveness through train-eval performance comparison.

Algorithm~\ref{alg:adaptive_threshold} presents the selective training control mechanism. The system maintains two moving averages of acceptance rates at different timescales: a short-term average ($\bar{\alpha}_{\text{short}}$) that responds quickly to recent changes, and a long-term average ($\bar{\alpha}_{\text{long}}$) that captures overall trends. Both averages are computed using exponential moving average (EMA):
\vspace{2pt}
\begin{equation}
\bar{\alpha}_t = \lambda \cdot \bar{\alpha}_{t-1} + (1 - \lambda) \cdot \alpha_t
\end{equation}
\vspace{2pt}
where $\lambda$ is the decay rate controlling the average's responsiveness.

This mechanism exhibits the desired self-regulating behavior. Initially, hidden state collection is enabled so that the draft model can adapt to the initial workload. Once sufficient data is collected, the system fine-tunes the draft model. If the fine-tuned model achieves higher acceptance rates than the previous model, the updated model is deployed. Otherwise, hidden state collection is disabled until a distribution shift is detected, at which point collection is automatically re-enabled.

\section{Evaluation}

\subsection{Experimental Setup}
\textbf{Implementation.} TIDE's inference serving engine is built on SGLang~\cite{zheng2024sglangefficientexecutionstructured}, while the draft model training engine is based on SpecForge~\cite{specforge2025}.

\textbf{Hardware.} We conduct inference experiments on a single node with up to eight NVIDIA H100 GPUs and draft model training on a separate node with four AMD Instinct MI250 GPUs. The nodes use a VAST shared-storage cluster to exchange hidden states and updated draft model checkpoints. The storage is mounted over NFSv3-RDMA through a 200~Gb/s InfiniBand link.

\textbf{Target Models.} We evaluate TIDE on four target models: gpt-oss-120b \cite{openai2025gptoss120bgptoss20bmodel}, Qwen3-235B-A22B \cite{qwen3technicalreport}, Llama-4-Scout-17B-16E \cite{meta2025llama4scout}, and Llama-3.3-70B-Instruct \cite{grattafiori2024llama3herdmodels}. These models span different architectures and parameter scales, providing a comprehensive evaluation across diverse model characteristics.

\textbf{Draft Model Configuration.}
We fix the number of candidate tokens to 3 across all experiments, as this empirically provides the best performance. 
The draft models use the same vocabulary as their corresponding target models. Since the target models are large enough to require tensor parallelism for efficient serving, the vocabulary size overhead in the draft models is negligible.

\textbf{Datasets.} We evaluate TIDE on diverse datasets spanning multiple domains: ShareGPT (conversational) \cite{aeala2023sharegpt}, Science (scientific text) \cite{camelai2023biology, camelai2023chemistry, camelai2023physics}, NuminaMath (mathematical reasoning) \cite{aimo2024numinamath}, and EvolCodeAlpaca (code generation) \cite{theblackcat2023evolcodealpaca}. Additionally, we evaluate on multilingual Alpaca datasets (Korean, Arabic, Chinese, French) \cite{freedomintelligence2023alpacakorean, freedomintelligence2023alpacaarabic, freedomintelligence2023alpacachinese, freedomintelligence2023alpacafrench} to assess TIDE's robustness to distribution shift, as vocabulary differences between languages represent the most significant source of distribution shift in our experiments.

\begin{figure}[t]
  \centering
  \includegraphics[width=1.03\columnwidth]{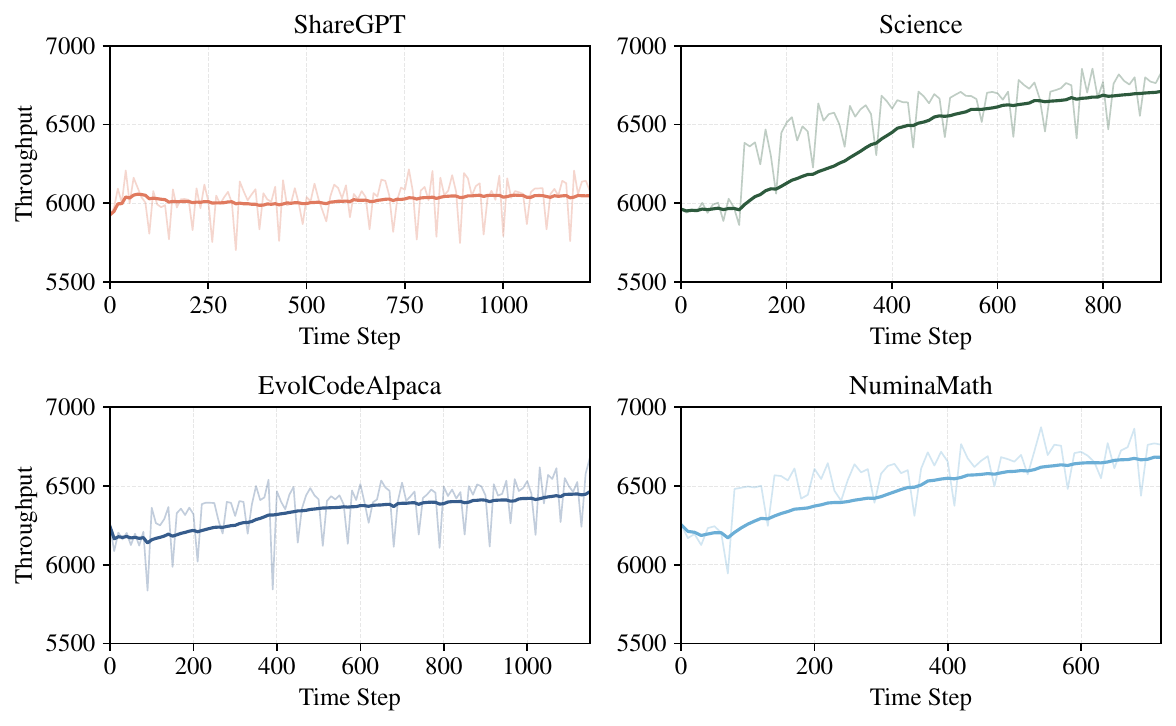}
  \caption{Throughput evolution over time across four datasets during inference serving using gpt-oss-120b as the target model. Each time step corresponds to 30 seconds of training time.}
  \label{fig:dataset_throughput}
\end{figure}

\begin{figure}[t]
  \centering
  \includegraphics[width=1.03\columnwidth]{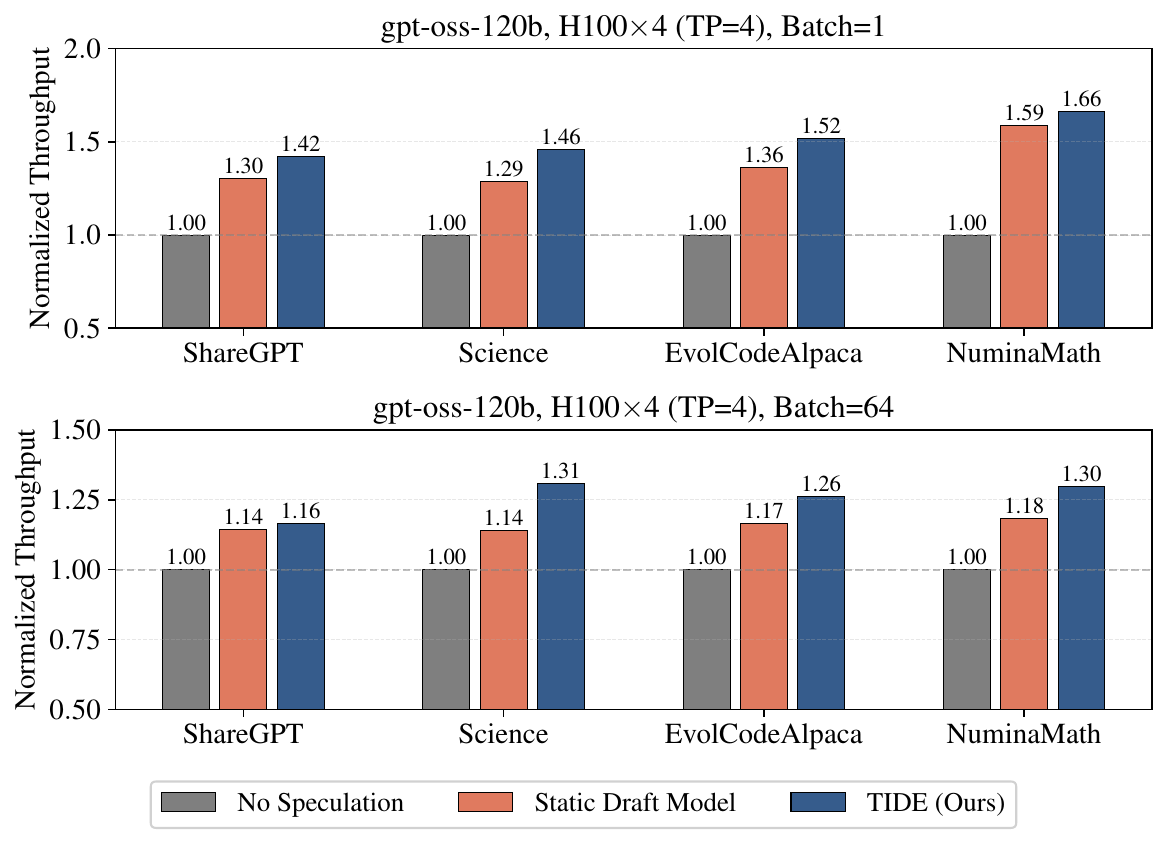}
  \caption{Normalized throughput comparison across four datasets at batch sizes 1 and 64. All values are normalized to the no-speculation baseline. At batch size 1, where speculative decoding is most effective, TIDE achieves up to 1.66$\times$ throughput over no speculation. At batch size 64, TIDE still achieves up to 1.31$\times$ improvement.}
  \label{fig:spec_decode_batch64}
\end{figure}

\subsection{Throughput Improvement over Time}
\label{subsection:throughput_improvement_over_time}
\cref{fig:dataset_throughput} shows the throughput evolution of TIDE during inference serving across four datasets. For Science, NuminaMath, and EvolCodeAlpaca, throughput consistently improves over time as the draft model is incrementally adapted using inference-time training signals. As shown in \cref{fig:spec_decode_batch64}, the improvement is most pronounced in latency-sensitive scenarios (batch size 1): TIDE achieves 1.46$\times$, 1.66$\times$, and 1.52$\times$ throughput over the no-speculation baseline, respectively, compared to 1.29$\times$, 1.59$\times$, and 1.36$\times$ for the static draft model. This is because at low batch sizes, GPU compute is underutilized during standard autoregressive decoding, allowing draft token verification to proceed with minimal overhead---and each accepted draft token directly reduces end-to-end latency. At batch size 64, where the GPU is more saturated, TIDE still achieves 1.31$\times$, 1.30$\times$, and 1.26$\times$ throughput over the baseline, compared to 1.14$\times$, 1.18$\times$, and 1.17$\times$ for the static draft model.

For ShareGPT, the static draft model already achieves 1.30$\times$ and 1.14$\times$ throughput over no speculation at batch sizes 1 and 64, respectively, and TIDE's online adaptation provides further gains to 1.42$\times$ and 1.16$\times$. This reflects a fundamental ceiling of speculative decoding on high-entropy conversational workloads \cite{xiang2025servegenworkloadcharacterizationgeneration, wang2024burstgpt}, not a limitation of TIDE: any draft model---static or adaptive---faces the same constraint. Notably, TIDE's adaptive control remains valuable in such cases by automatically reducing unnecessary training when further adaptation yields diminishing returns.

In contrast, TIDE's impact is most pronounced when draft--target alignment is poor. As shown in \cref{fig:spec_decode_comparison}, when a static draft model trained on English data is applied to non-English workloads, speculative decoding degrades throughput to 0.70--0.86$\times$ of the no-speculation baseline. TIDE recovers performance to 1.08--1.11$\times$, demonstrating its ability to adapt to workloads that static approaches cannot serve effectively. Taken together, these results illustrate that speculative decoding does not uniformly accelerate all workloads---its effectiveness varies significantly across domains, ranging from modest gains on high-entropy conversational data to substantial speedups on structured domains such as mathematics and code. Pre-training a draft model to cover all possible workloads is impractical: production workloads span an open-ended set of domains, and even if broad coverage were achievable at training time, the distribution of incoming requests shifts over time as user behavior and application demands evolve. This workload-dependent and temporally non-stationary nature of speculative decoding effectiveness highlights the need for runtime adaptation logic that determines when to enable speculative decoding and when to trigger draft model training.

\subsection{Adaptive Speculative Decoding Control}
\begin{figure*}[t]
  \centering
  \includegraphics[width=\textwidth]{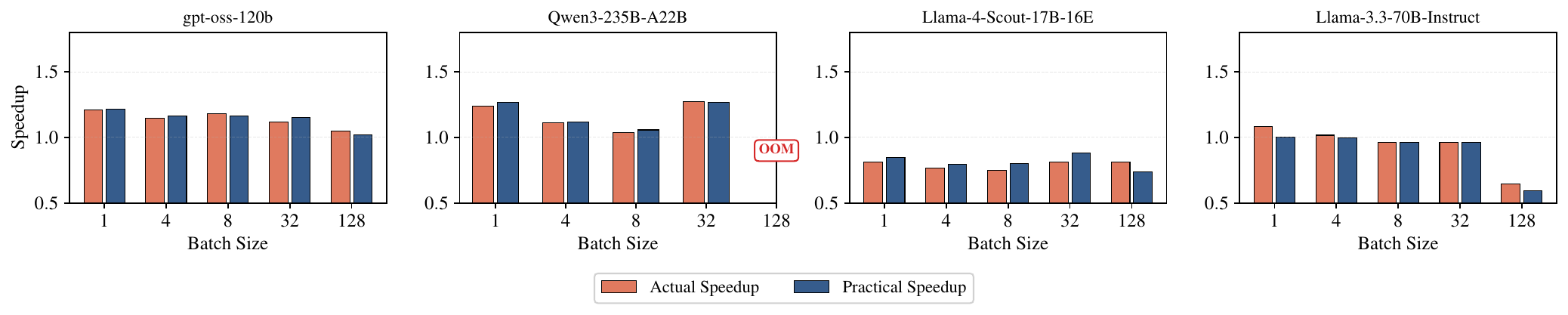}
  \caption{Comparison of practical and actual speedup across batch sizes. Actual speedup is measured using four H100 GPUs.}
  \label{fig:practical_speedup}
\end{figure*}

\begin{figure*}[t]
  \centering
  \includegraphics[width=\textwidth]{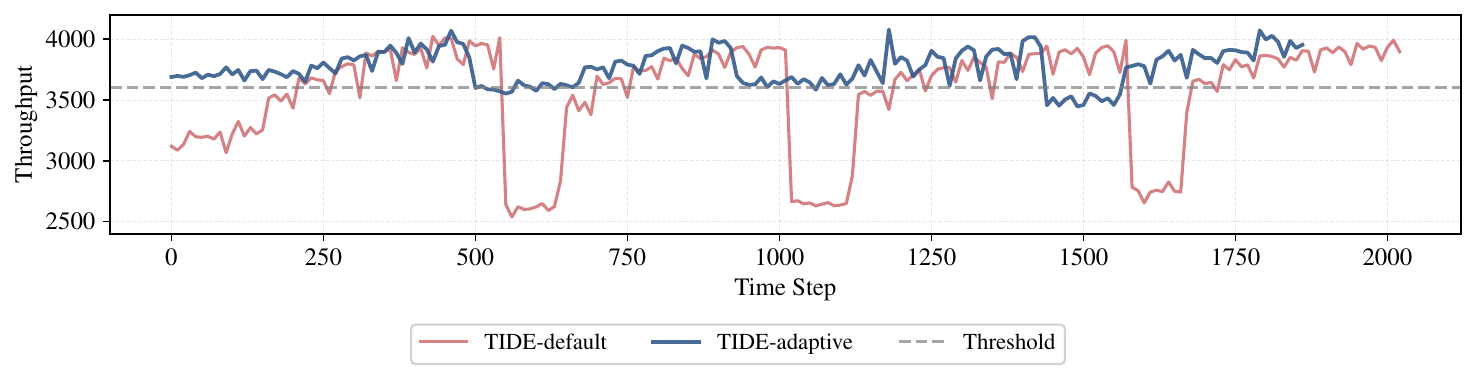}
  \caption{Throughput over time for TIDE-default and TIDE-adaptive under distribution shifts. Sequential language transitions (Korean → Arabic → Chinese → French) using Alpaca datasets demonstrate adaptive performance. Both experiments process an identical total workload, while TIDE-adaptive finishes at an earlier timestep.}
  \label{fig:adaptive_control}
\end{figure*}
\begin{figure}[t]
\centering
\includegraphics[width=1.03\columnwidth]{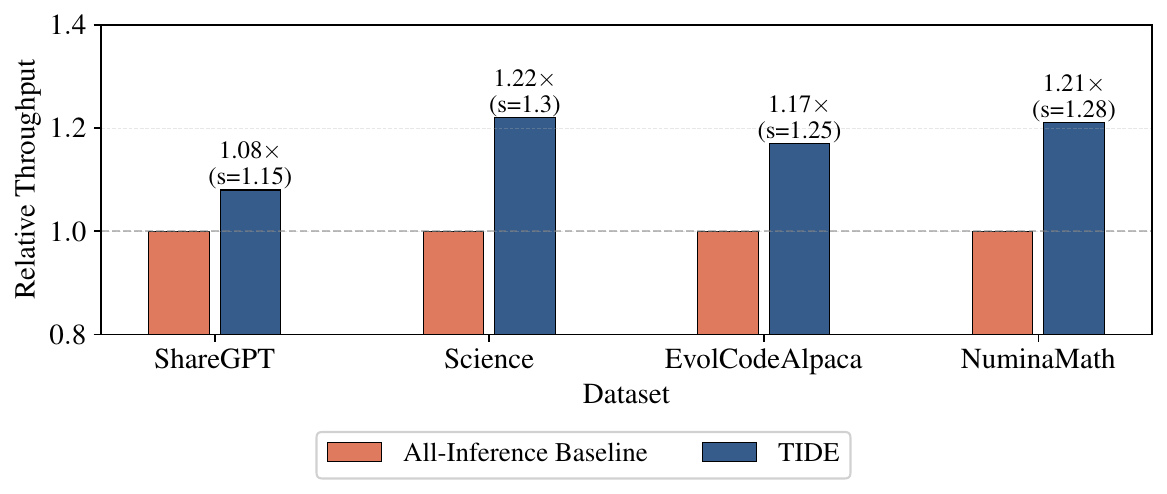}
\caption{End-to-end throughput of two allocations using the same hardware budget (8$\times$ H100 + 4$\times$ MI250), normalized to All-Inference. All-Inference assigns both GPU nodes to standard inference with speculative decoding disabled; TIDE uses the H100 node for speculative decoding with the target and draft models, while the MI250 node concurrently trains the draft model on the same live workload. Parentheses indicate the resulting speculative-decoding speedup ($s$) on the H100 serving node.}
\label{fig:hetero_dataset}
\end{figure}

We validate our analytical model for predicting speculative decoding speedup and use it to implement adaptive control mechanisms. \cref{fig:practical_speedup} compares practical speedup predicted by our analytical model (Equation~\ref{eq:practical_speedup}) against actual measured speedup across different batch sizes. Across all four target models, the practical speedup predicted by our analytical model closely tracks the actual measured speedup within 9\% error, validating the accuracy of our model for adaptive control decisions.

Using this model, we calculate the minimum acceptance length required for performance improvement at each batch size. We evaluate two configurations: (1) TIDE-default, which always enables speculative decoding regardless of batch size and acceptance length, and (2) TIDE-adaptive, which dynamically enables or disables speculative decoding based on whether acceptance length exceeds the calculated threshold. \cref{fig:adaptive_control} compares throughput over time for both configurations. TIDE-adaptive mitigates throughput degradation during distribution shifts by adaptively disabling speculative decoding when acceptance length is insufficient, while TIDE-default experiences more severe performance drops.

\begin{figure*}[!ht]
  \centering
  \includegraphics[width=\textwidth]{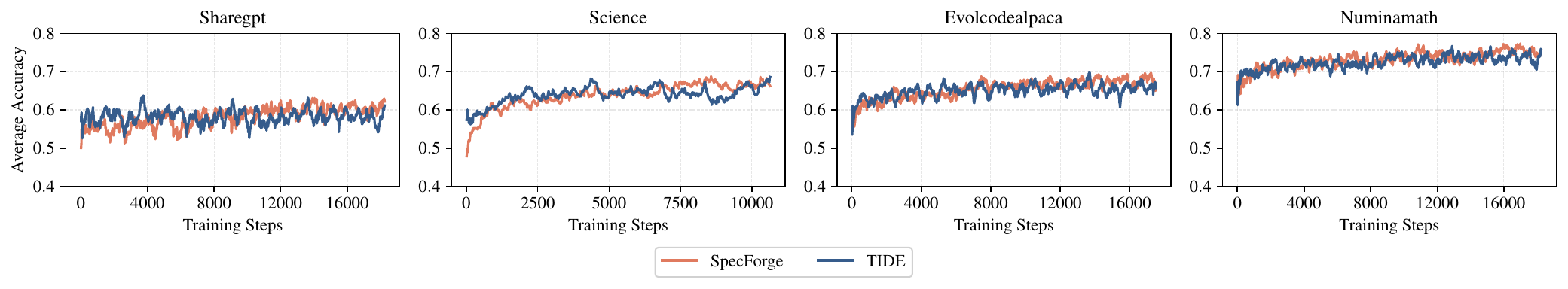}
  \caption{Draft model training accuracy comparison between SpecForge offline and TIDE across four datasets using a global batch size of 16 on four MI250 GPUs. Accuracy measures the top-1 token prediction match rate between the draft model and the target model.}
  \label{fig:training_accuracy}
\end{figure*}

\subsection{Training Efficiency Comparison}
\label{subsection:training_efficiency_comparison}
We compare TIDE against two state-of-the-art training approaches for EAGLE-3 draft models in SpecForge: offline training and online training.

\textbf{SpecForge Offline Training.} This approach assumes a pre-collected dataset of input prompts paired with their corresponding target model-generated outputs. It first concatenates each input--output pair into a single sequence and runs prefill through the target model to extract intermediate hidden states, which are then stored to disk. Since draft model training typically requires multiple passes over the dataset, storing hidden states avoids reloading the target model and recomputing prefill at every epoch. However, this comes at the cost of substantial storage as shown in Table~\ref{tab:storage_comparison}.

\textbf{SpecForge Online Training.} To reduce disk usage, this approach regenerates hidden states on-demand at every epoch instead of storing them. However, this requires reloading the target model and recomputing prefill for every pass over the dataset, resulting in significantly slower training time as shown in Table~\ref{tab:training_time}.

In contrast to both approaches, TIDE extracts hidden states directly from the target model's inference step, eliminating the dedicated prefill phase that accounts for 40--67\% of total training time in SpecForge methods (Table~\ref{tab:training_time}). Like SpecForge offline, TIDE stores hidden states to disk, but only maintains a small buffer that is consumed and discarded after each training cycle rather than retaining the entire dataset. This drastically reduces storage requirements---by up to 24$\times$ compared to offline training (e.g., 4.66\,TB to 0.19\,TB for gpt-oss-120b) as shown in Table~\ref{tab:storage_comparison}---while also enabling faster adaptation: as shown in Table~\ref{tab:training_time}, TIDE is 1.67$\times$ faster than SpecForge offline and 3.02$\times$ faster than SpecForge online on the ShareGPT dataset with gpt-oss-120b. Crucially, these efficiency gains do not sacrifice training quality: \cref{fig:training_accuracy} confirms that TIDE achieves comparable final accuracy to both SpecForge methods across four datasets, while enabling continuous adaptation during live serving that existing approaches cannot provide.

\begin{table}[t]
\centering
\caption{Storage requirements for hidden states across different target models. SpecForge offline stores all hidden states for the entire dataset, while TIDE only maintains a small buffer for active training batches.}
\label{tab:storage_comparison}
\begin{small}
\begin{tabular}{lcc}
\toprule
\textbf{Target Model} & \textbf{SpecForge offline} & \textbf{TIDE} \\
\midrule
gpt-oss-120b & 4.66 TB & 0.19 TB \\
Qwen3-235B-A22B & 6.63 TB & 0.27 TB \\
Llama-4-Scout-17B-16E & 8.29 TB & 0.34 TB \\
Llama-3.3-70B-Instruct & 13.26 TB & 0.55 TB \\
\bottomrule
\end{tabular}
\end{small}
\end{table}

\begin{table}[t]
\centering
\caption{Training time comparison for gpt-oss-120b on ShareGPT (100k) dataset. TIDE eliminates prefill overhead by reusing hidden states from inference serving.}
\label{tab:training_time}
\begin{footnotesize}
\begin{tabular}{lcccc}
\toprule
\textbf{Method} & \textbf{Prefill} & \textbf{Train} & \textbf{Total} & \textbf{Speedup} \\
\midrule
SpecForge offline & 6.16 h & 9.16 h & 15.32 h & 1.00$\times$ \\
SpecForge online & 18.48 h & 9.16 h & 27.64 h & 0.55$\times$ \\
TIDE & - & 9.16 h & \textbf{9.16 h} & \textbf{1.67$\times$} \\
\bottomrule
\end{tabular}
\end{footnotesize}
\end{table}

\subsection{Shared-Storage I/O Scalability}
We further quantify whether storage writes can keep pace under multi-node contention. For gpt-oss-120b, EAGLE-3 stores BF16 hidden states from three target-model layers per token. Given a hidden size of 2,880 and 2 bytes per BF16 element, each token produces $2{,}880 \times 3 \times 2 = 17{,}280$ bytes, or approximately 16.9~KiB, of training signals. Each inference serving instance uses tensor parallelism across four GPUs (TP$=4$) and generates approximately 6,000--7,000 tokens/s, resulting in 99--116~MiB/s of write traffic.

Our VAST storage sustains 564~MB/s with one writer and 1.47~GB/s with eight concurrent serving instances across four nodes. The latter corresponds to approximately 184~MB/s per instance, providing 1.6--1.9$\times$ headroom over the observed demand. The aggregate traffic also uses only a small fraction of the 200~Gb/s InfiniBand link's approximately 25~GB/s capacity. Although these measurements characterize storage capacity rather than per-operation latency, they demonstrate that shared-storage I/O is not a bottleneck at the evaluated multi-node scale.

\subsection{Heterogeneous GPU Allocation}
\begin{figure}[t]
  \centering
  \includegraphics[width=1.03\columnwidth]{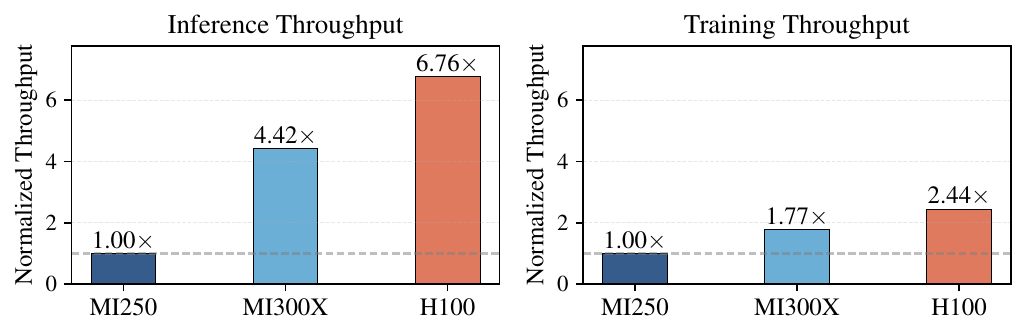}
  \caption{Per-GPU throughput comparison for inference and draft model training, normalized to MI250 baseline. Inference throughput measured on gpt-oss-120b using SGLang with tensor parallelism. Training throughput was measured using PyTorch with FSDP on single nodes equipped with 4 MI250, 8 MI300X, and 8 H100 GPUs, respectively.}
  \label{fig:gpu_throughput}
\end{figure}

As modern datacenters incrementally adopt new accelerator generations, heterogeneous GPU clusters are inevitable. We evaluate how TIDE leverages this heterogeneity by profiling inference and training throughput across GPU types. \cref{fig:gpu_throughput} shows that the performance gap across GPU generations is significantly larger for inference than for training: compared to MI250, H100 achieves 6.76$\times$ higher inference throughput but only 2.44$\times$ higher training throughput, with MI300X showing a similar pattern (4.42$\times$ vs.\ 1.77$\times$). This asymmetry means that dedicating high-end GPUs to inference---where they provide the greatest advantage---and assigning older GPUs to draft model training---where the relative performance penalty is small---maximizes overall system efficiency. Importantly, since TIDE fully decouples inference and training onto separate nodes that communicate only through shared storage for hidden states and model checkpoints, this heterogeneous deployment requires no cross-vendor driver compatibility, no GPU-to-GPU interconnect, and no modifications to existing datacenter infrastructure.

To quantify the benefits of this approach, we compare two allocation strategies under the same hardware budget of a single H100 node with 8 GPUs and a single MI250 node with 4 GPUs. The All-Inference baseline assigns both the H100 and MI250 nodes to standard target-model inference with speculative decoding disabled. TIDE instead uses the H100 node for speculative decoding with the target and draft models, while assigning the MI250 node to asynchronous draft model training.

\cref{fig:hetero_dataset} shows that, once draft model training has converged, TIDE achieves 1.08--1.22$\times$ throughput improvement over the all-inference baseline. Although TIDE uses the MI250 node for training instead of inference, it still achieves higher total throughput. This shows that the throughput gained from speculative decoding on the H100 node exceeds the throughput that the MI250 node would provide through direct inference. The improvement correlates with the speculative decoding speedup achieved through draft model training, ranging from $s=1.15$ (ShareGPT, 1.08$\times$ throughput) to $s=1.30$ (Science, 1.22$\times$ throughput). These variations reflect differences in output distribution characteristics and draft model learning difficulty across datasets. For instance, Science dataset's more structured output enables better draft model learning, resulting in higher acceptance rates and greater speedup. This result demonstrates that TIDE's benefits vary with dataset characteristics and highlights the importance of considering workload properties when deploying heterogeneous training strategies.

\begin{table}[t]
\centering
\caption{Dataset sizes and aggregate input and output sequence lengths. Lengths are reported in millions of tokens.}
\label{tab:dataset_sequence_lengths}
\begin{small}
\begin{tabular}{lcc}
\toprule
\textbf{Dataset} & \makecell{\textbf{Total Input}\\\textbf{Length (M)}} & \makecell{\textbf{Total Output}\\\textbf{Length (M)}} \\
\midrule
ShareGPT & 106.18 & 512.65 \\
Science & 7.98 & 286.15 \\
EvolCodeAlpaca & 50.77 & 491.84 \\
NuminaMath & 21.07 & 234.18 \\
\bottomrule
\end{tabular}
\end{small}
\end{table}

Prefill/decode disaggregation provides another possible allocation, with the MI250 node serving as a prefill worker and the H100 node performing decoding. However, the workloads in \cref{fig:hetero_dataset} are decode-heavy. As shown in \cref{tab:dataset_sequence_lengths}, output sequences are longer than input sequences for all four datasets; the geometric mean of the per-dataset output-to-input length ratios is 11.7$\times$. Direct profiling on H100 also shows that prefill accounts for at most 1\% of end-to-end inference time. Assigning the MI250 node exclusively to prefill would therefore leave it substantially underutilized for the decode-heavy workloads evaluated here.

\section{Discussion}

While TIDE demonstrates consistent throughput improvements across diverse workloads, several limitations merit discussion.

\subsection{Draft Architecture Dependence}
TIDE's training signal extraction is tightly coupled to the EAGLE-3 draft architecture, which predicts next tokens from the target model's intermediate hidden states at selected decoder layers. This design choice avoids additional target-model computation for training-signal generation but also limits generality: draft architectures that do not consume target model hidden states as input---such as independent draft models trained via standard distillation \cite{liu2024onlinespeculativedecoding}---cannot directly benefit from TIDE's training signal extraction pipeline. Extending TIDE to support alternative draft architectures would require rethinking how training signals are collected and what intermediate representations are reused, which we leave as future work.

\subsection{Adaptation Latency Under Rapid Distribution Shifts}
TIDE detects distribution shifts by comparing short-term and long-term exponential moving averages of acceptance rates. While this mechanism effectively identifies gradual workload transitions, it introduces inherent latency in responding to abrupt distribution changes. Upon detecting a shift, TIDE must still collect sufficient training signals, transfer them to the training engine, fine-tune the draft model, and deploy the updated checkpoint. If the workload shifts faster than this adaptation cycle, the temporal locality window may close before the updated draft model is deployed. 

TIDE's adaptive control mechanism provides a short-term mitigation: upon detecting a distribution shift, the controller can promptly disable speculative decoding, ensuring that inference throughput does not degrade below the baseline of standard autoregressive generation. While this prevents performance regression, it is a conservative fallback that forgoes the potential benefits of speculative decoding during the adaptation period. A more fundamental solution would be to maintain a pool of domain-specialized draft model checkpoints, allowing the system to switch to a previously trained checkpoint upon detecting a familiar distribution rather than retraining from scratch. This approach would substantially reduce adaptation latency for recurring workloads, which we discuss further in \cref{subsec:catastrophic_forgetting}.

\subsection{Catastrophic Forgetting Under Recurring Workloads}
\label{subsec:catastrophic_forgetting}
TIDE continuously fine-tunes the draft model on recent inference data to exploit short-term temporal locality. However, this incremental adaptation inherently favors the current workload distribution and may degrade performance on previously learned patterns. We have not evaluated how such degradation accumulates over repeated workload transitions, leaving long-term behavior under recurring workloads as an open question. In production environments where workloads cycle through recurring patterns---for example, code generation during business hours followed by conversational queries in the evening---the draft model may repeatedly lose and relearn the same distributions.

A natural extension is to maintain a pool of domain-specialized draft model checkpoints, where each checkpoint is associated with a detected workload distribution. Upon a distribution shift, the system can select the closest matching checkpoint from the pool and resume fine-tuning from it, rather than adapting a misaligned model from scratch. Importantly, TIDE's core contribution---runtime training signal extraction from target model inference---is orthogonal to the draft model management strategy. The same serving-time training signal extraction pipeline can serve multiple draft models simultaneously, and the adaptive control mechanisms can be extended to route requests to the most appropriate checkpoint. We leave the design of such a multi-draft-model management strategy as future work.

\section{Conclusion}
We present TIDE, a serving-engine-native framework that continuously adapts draft models at runtime by reusing intermediate hidden states from target model inference without additional target-model computation and with no additional serving-time overhead for training-signal extraction. Combined with adaptive control that dynamically toggles speculative decoding and training, and a decoupled architecture that leverages heterogeneous GPU clusters, TIDE consistently improves throughput across diverse workloads---especially in latency-sensitive scenarios and on out-of-distribution workloads where static draft models fail. Future work includes extending TIDE with a pool of domain-specialized draft model checkpoints to mitigate catastrophic forgetting under recurring workloads, and generalizing the training signal extraction pipeline to draft architectures beyond EAGLE-3.

\section{Acknowledgment}

This work was supported by Institute of Information \& Communications Technology Planning \& Evaluation (IITP) grant funded by the Korea government (MSIT) (No. RS-2025-02304554, Efficient and Scalable Framework for AI Heterogeneous Cluster Systems).

\bibliographystyle{IEEEtran}
\bibliography{main}

\onecolumn

\end{document}